\documentclass[letterpaper]{article} 
\usepackage[preprint]{aaai2027}  
\usepackage[hyphens]{url}  
\usepackage{graphicx} 
\usepackage{natbib}  
\usepackage{caption} 
\usepackage{amsmath,amssymb}
\usepackage{booktabs}
\usepackage{array}
\usepackage{dblfloatfix} 

\DeclareCaptionStyle{ruled}{labelfont=normalfont,labelsep=colon,strut=off}

\title{
MAP-Graph: Provenance-Aware Shared Memory for Multi-Agent Workflows}

\author{
Yiqi Wang\textsuperscript{*},\quad
Zihao Yan,\quad
Jiaqi Zhang,\quad
Zhangkai Wu,\\
Mingkai Zheng,\quad
Zequn Sun,\quad
Yanming Zhu,\quad
Taotao Cai
}

\affiliations{
\textsuperscript{*}
\texttt{yiqi.wang.jennie@gmail.com}
}

\begin{document}

\maketitle

\begin{abstract}
Shared memory helps language-model agents reuse information across long workflows, yet relevant evidence may not be admissible for a particular agent or action. Because restrictions propagate through derivations, summaries can conceal private, poisoned, untrusted, or revoked sources, enabling unauthorized reads or unsafe actions. Existing approaches provide semantic retrieval, scoped access, or lineage tracking, but do not clearly separate hard authorization from graded trust or adapt evidence requirements to action risk.
We introduce MAP-Graph, a provenance-aware memory layer that represents agents, sources, memories, claims, and actions in a typed execution graph. It traces ancestry, excludes permission-ineligible records, reranks eligible memories by semantic similarity and multiplicative path trust, and applies a risk-sensitive gate before action execution while retaining affected lineage for audit. On a controlled benchmark of 2,700 synthetic tasks per method across three domains, MAP-Graph achieves 94.96\% overall task success, 72.70\% exact decision accuracy, and 90.22\% in the clean setting, where success requires a correct \textsc{Allow} rather than a safe intervention. 
Ablations isolate the roles of permission filtering, path trust, and action gating, while transfer tests with two additional backbones preserve the exact-decision and access-control advantages. These results support provenance as an operational control signal, rather than only post-hoc audit metadata, within the evaluated setting.

\end{abstract}

\section{Introduction}
\label{sec:introduction}

\begin{figure*}[!t]
    \centering
    \includegraphics[
        width=0.98\textwidth,
        height=0.48\textheight,
        keepaspectratio
    ]{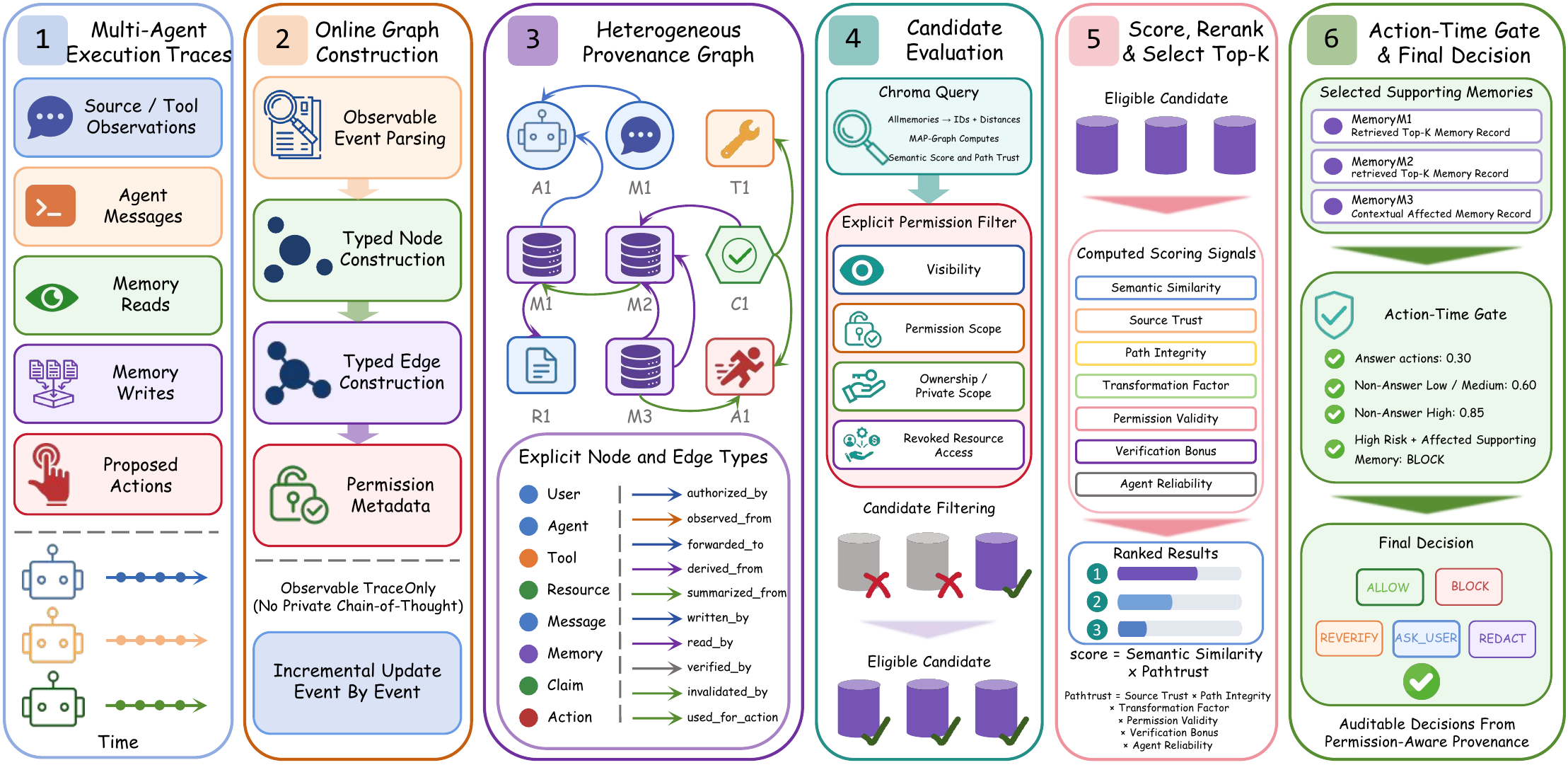}
    \caption{Overview of MAP-Graph. A task-scoped typed provenance graph records memory origins and downstream dependencies. MAP-Graph combines semantic retrieval, recursive ancestry tracing, permission filtering, path-trust reranking, and risk-sensitive action gating to return \textsc{Allow}, \textsc{Block}, \textsc{Reverify}, \textsc{Redact}, or \textsc{AskUser}.}
    \label{fig:overview-map}
\end{figure*}

Language-model agents increasingly plan, use tools, and reason over long horizons~\cite{yao2023react,shinn2023reflexion}; multi-agent frameworks add role specialization and collaborative decomposition~\cite{park2023generativeagents,li2023camel,wu2024autogen,hong2024metagpt}. 
Shared memory preserves intermediate knowledge across agents and sessions~\cite{packer2023memgpt,zhong2024memorybank,wang2023longmem,zhang2025memorysurvey}. As workflows incorporate more agents, tools, and external resources, this memory becomes execution state: one agent's record can later determine another agent's answer or tool action~\cite{gao2024memorysharing,rezazadeh2025collaborativememory,zhang2025gmemory,wang2025mirix,margalit2026governed}.

This creates a failure mode that retrieval quality alone cannot diagnose. Vector retrieval can return the most relevant record~\cite{lewis2020rag,guu2020realm,karpukhin2020dpr} while remaining blind to whether that record may be read or used. Moreover, derivation can hide the relevant restriction: a summary may omit the fact that its source was private, poisoned, untrusted, or subsequently revoked. The derived memory can therefore remain highly relevant while enabling unauthorized disclosure or an unsafe downstream action~\cite{chao2026stale,dash2026memorypoisoning,pulipaka2026sleeper,wang2026mempoison}. 

Existing approaches cover substantial pieces of this problem. Private/shared memory tiers enforce local access policies; graph-structured memory represents relationships~\cite{edge2024graphrag,guo2025lightrag,gutierrez2024hipporag}; and provenance models record  derivation~\cite{belhajjame2013provdm,moreau2013prov}. Recent governed shared memory further combines scoped retrieval, provenance tracking, temporal supersession, and policy propagation in a production service~\cite{margalit2026governed}. That study intentionally evaluates a live system rather than comparative action policies, leaving open how hard authorization, graded ancestry trust, and action-risk gating interact under common controlled conditions. More generally, local metadata need not expose inherited restrictions, representation alone does not determine authorization, and a retrieval-time decision need not match the risk of the eventual action. The unresolved question is therefore: \emph{how can a shared-memory system retrieve useful records while enforcing agent-, ancestry-, and action-specific admissibility?}


Our answer begins with one distinction: \emph{retrieval is not authorization}. \emph{Semantic relevance} asks whether a memory is useful for the query; \emph{provenance admissibility} asks whether the requesting agent may use it for the proposed action. 
Operationalizing the distinction raises three challenges.
\textbf{(C1) Recursive Ancestry.} 
Admissibility may depend on multi-step derivation chains across sources and derived memories, so inspecting only a record's local metadata or immediate source is insufficient.
\textbf{(C2) Heterogeneous Constraints.} 
Permission and visibility determine whether a record is eligible for retrieval, whereas source reliability, verification, and other provenance-quality signals should grade the eligible evidence without allowing semantic similarity to override a hard access restriction. 
\textbf{(C3) Action-Dependent Admissibility.} 
Evidence sufficient for a low-risk response may be inadequate for a consequential action, so admissibility cannot be decided only once at retrieval time.

We propose \textbf{MAP-Graph}, a provenance-aware memory layer that addresses C1-C3 in sequence.
A typed execution graph records agents, sources, memories, claims, actions, and their derivations. 
Retrieval first applies permission eligibility, then combines semantic relevance with multiplicative path trust computed from recorded ancestry. Once an action is proposed, a risk-sensitive gate re-evaluates its supporting memories. Affected descendants remain marked rather than deleted, preserving evidence for audit while constraining downstream use.
Figure~\ref{fig:overview-map} summarizes the workflow.

We evaluate seven baselines and six ablations on 2,700 synthetic tasks across three domains, plus three backbones on a stratified 540-task subset. MAP-Graph reaches 94.96\% task success, 72.70\% exact accuracy, and 90.22\% clean success. Removing its hard permission filter improves utility but allows every observed unauthorized read, a failure hidden by aggregate utility and leakage alone. These are single-run controlled results, not deployment-scale claims.

In summary, this paper makes the following contributions:
\begin{enumerate}
    \item We formulate provenance-aware shared-memory use as a controlled decision problem that separates semantic relevance from agent- and action-specific provenance admissibility.

    \item We propose \textbf{MAP-Graph}, which combines typed provenance representation, permission filtering, ancestry-aware trust reasoning, risk-sensitive action gating, and audit-preserving descendant marking.

    \item We provide a three-domain controlled benchmark with seven baselines, six ablations, and a stratified three-backbone comparison.
\end{enumerate}
\section{Background and Problem Formulation}
\label{sec:background-problem}

\subsection{Multi-Agent Shared Memory}
\label{sec:multi-agent-shared-memory}

We consider a set of agents $\mathcal{A}=\{a_1,\ldots,a_n\}$ collaborating on a long-horizon task. At each step an agent may observe information, call a tool, exchange messages, generate claims, or make partial decisions, producing information worth storing for later reuse. A shared memory store $\mathcal{M}$ maintains it across agents and time, with a memory entry denoted
\begin{equation}
m_i = (\mathrm{id}_i,c_i,o_i,v_i,d_i,\ell_i,\eta_i),
\label{eq:memory-entry}
\end{equation}
where $\mathrm{id}_i$ is the identifier, $c_i$ the content, $o_i$ the owner agent, $v_i$ the visibility, $d_i$ the source and parent-memory references, $\ell_i$ an optional trust label, and $\eta_i$ additional metadata. Final decisions are represented separately as Action nodes rather than as memory entries.

A score computed from $c_i$ alone cannot reveal who may read the record, which source it derives from, or whether an ancestor has been marked private, untrusted, or revoked, so the store may return an on-topic record that should not support the current action.

\subsection{Memory Provenance}
\label{sec:memory-provenance}


Memory provenance records a record's origin and derivation. MAP-Graph uses it operationally: recorded ownership and inherited permission scope determine eligibility, while source and derivation metadata inform path trust and action gating. The benchmark supplies explicit visibility, ownership, trust, risk, and revocation fields; MAP-Graph records their lineage and does not infer a universal notion of truth from free text. Its registered entities and relations are specified in Section~\ref{sec:graph-schema} and Appendix~\ref{app:schema}.

\subsection{Provenance-Aware Retrieval Objective}
\label{sec:provenance-aware-retrieval-objective}

Let $\mathcal{M}_t$ denote the records created or seeded for task $t$, and $\mathrm{CanRead}(a,m)\in\{0,1\}$ the implemented permission check for agent $a$. For a query $q$, MAP-Graph first restricts retrieval to
\begin{equation}
\mathcal{C}(a)=\{m\in\mathcal{M}_t:\mathrm{CanRead}(a,m)=1\},
\label{eq:permission-candidates}
\end{equation}
then scores each eligible $m$ by a semantic score $s(m,q)$ from cosine distance and a path-trust value $\rho(m,a)$ computed from recorded ancestors:
\begin{equation}
\mathrm{Score}(m,q,a)=s(m,q)\,\rho(m,a).
\label{eq:implemented-retrieval-objective}
\end{equation}
The system returns the five highest-scoring candidates. This is a deterministic rule over explicit metadata, not a learned calibration of factual correctness.

The factorization is not merely notational: separating the binary gate $\mathrm{CanRead}$ from the graded factor $\rho$, and both from the content score $s$, fixes what an implementation must supply. Three requirements follow directly, restating the challenges of Section~\ref{sec:introduction} as properties of the objective rather than observations about prior systems. \textbf{Ancestry-aware evaluation (C1):} trust must be computed over the transitive recorded ancestry of $m$, while access restrictions must remain attached as memories are derived. In our implementation, a derived memory receives the intersection of its referenced scopes at write time; query-time $\mathrm{CanRead}$ then checks that stored scope and any directly referenced source. \textbf{Separation of hard and graded constraints (C2):} $\mathrm{CanRead}$ must filter the candidate set while $\rho$ multiplies the surviving scores---folding permission into $\rho$ as a large penalty would let a sufficiently similar record outrank the restriction, and folding trust into $\mathcal{C}(a)$ would discard usable evidence whose reliability is merely reduced. \textbf{A second, action-indexed decision (C3):} $\mathrm{Score}$ ranks records for a query but carries no notion of what the evidence will be used for, so admissibility for a proposed action requires a separate decision, taken after the action is known and parameterized by its risk. Sections~\ref{sec:graph-schema}--\ref{sec:trust-conflict-invalidation} present the mechanism satisfying each in turn.

\section{MAP-Graph Framework}
\label{sec:map-graph-framework}

\begin{figure*}[!t]
    \centering
    \includegraphics[
        width=0.98\textwidth,
        keepaspectratio
    ]{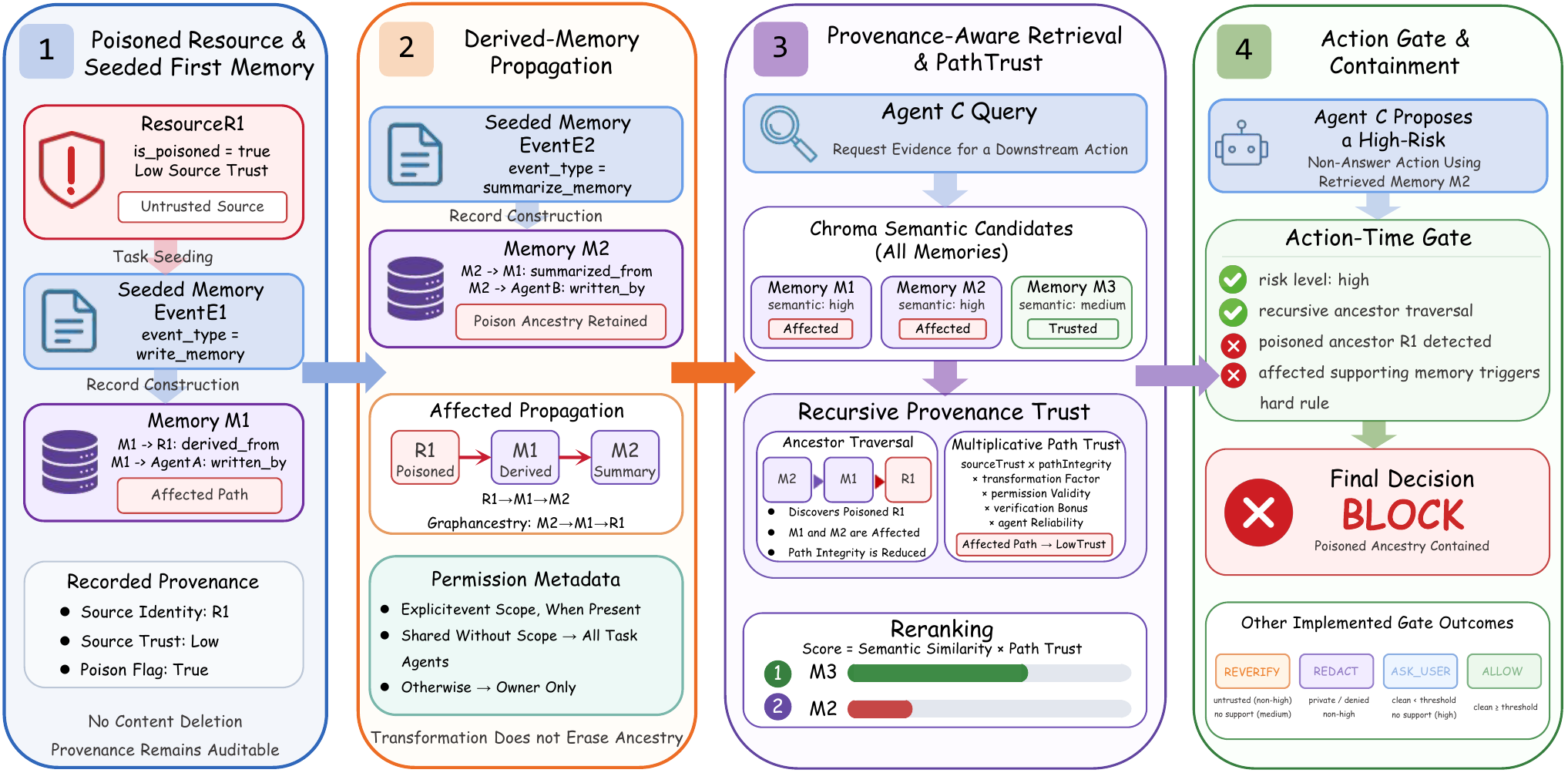}
    \caption{Running example of poisoned-memory propagation and containment in MAP-Graph. A poisoned resource $R_1$ produces memory $M_1$, which is summarized into $M_2$ while retaining its provenance ancestry. During retrieval, recursive ancestry tracing identifies the poisoned source and reduces the path trust of the affected memories. When $M_2$ supports a high-risk action, the action-time gate returns \textsc{Block}, while preserving the affected lineage for audit.}
    \label{fig:running-example}
\end{figure*}

MAP-Graph is a task-scoped memory backend placed between a multi-agent workflow and its proposed action. For every task, it resets a NetworkX \texttt{MultiDiGraph}, an ephemeral Chroma collection, and the associated source and memory catalogs. The graph records observable execution lineage, while the vector collection provides semantic candidates; the two representations are coupled through common memory identifiers.

Figure~\ref{fig:running-example} illustrates the complete processing path through a poisoned-memory example. A poisoned resource $R_1$ produces memory $M_1$, which is subsequently summarized into $M_2$. Although the transformation changes the memory content, it preserves the ancestry path $M_2 \rightarrow M_1 \rightarrow R_1$. When Agent~C requests evidence for a high-risk action, recursive provenance evaluation discovers the poisoned ancestor and reduces the path trust and final ranking scores of the affected memories. If the affected memory still supports the proposed action, the action-time gate returns \textsc{Block}, while retaining the corresponding lineage for audit.

The following subsections explain this process in order: the schema and construction rules make ancestry recoverable (C1), retrieval separates hard permission filtering from graded trust reranking (C2), and the action gate re-evaluates admissibility once the proposed action is known (C3).

\subsection{Graph Schema: Registered Node and Edge Types}
\label{sec:graph-schema}

MAP-Graph represents shared memory as a heterogeneous directed provenance graph $\mathcal{G} = (\mathcal{V}, \mathcal{E}, \tau_V, \tau_E)$, where $\tau_V$ is a partial map over registered node types and $\tau_E$ assigns edge types.

The registered vocabulary has eight node types (User, Agent, Tool, Resource, Message, Memory, Claim, Action) and ten edge types: \texttt{authorized\_by}, \texttt{observed\_from}, \texttt{forwarded\_to}, \texttt{derived\_from}, \texttt{summarized\_from}, \texttt{written\_by}, \texttt{read\_by}, \texttt{verified\_by}, \texttt{invalidated\_by}, and \texttt{used\_for\_action}. The derivation edges are what make ancestry recoverable from an export alone, the precondition for the C1 evaluation in Section~\ref{sec:provenance-aware-memory-retrieval}. Task identifier, domain, step, visibility, permission scope, trust labels, and risk tags are attributes rather than node types. The graph is thus an execution-lineage representation, not a semantic knowledge graph: no generic \texttt{Contradicts}, \texttt{Updates}, or temporal-order edges are created, and no automatic open-domain conflict resolution is claimed. Per-type definitions, serialized labels, and the constants entering Equation~\ref{eq:path-trust} are in Appendix~\ref{app:schema}.

\subsection{Memory Construction and Provenance Capture}
\label{sec:memory-construction-provenance-capture}

Construction uses only observable workflow state and task-provided metadata; no private chain-of-thought is required. At task initialization MAP-Graph registers the task-provided agent roles and source documents and inserts any benchmark-provided initial memory events. When an agent observes a source or tool output, the backend creates a record whose \texttt{derived\_from} field includes that identifier. When an agent writes a message-derived memory, the record inherits the identifiers of retrieved memories and visible sources, and its permission scope is the intersection of the referenced scopes; if no referenced scope exists all task agents are eligible, and if the intersection is empty access falls back to the writing agent.

Observable messages and up to three coarse Claim nodes are retained for graph inspection. The write path rejects empty text, while each non-empty write creates a distinct embedded Memory node with its own lineage. Semantically equivalent entries are not merged, and new observations do not mutate or supersede prior text; no generic supersession or conflict-resolution state machine is implemented.

An explicit \texttt{revoke\_permission} event changes the referenced source to revoked visibility, clears its permission scope, and immediately recomputes records that directly reference it or already list it as an affected ancestor, adding \texttt{invalidated\_by} edges where applicable. Other descendants detect the revoked ancestor when recursive trust is next evaluated, and task finalization marks them in the audit graph. ``Invalidation'' below therefore refers to this explicit revocation path, not a lifecycle inferred from natural-language contradictions.

\subsection{Provenance-Aware Memory Retrieval}
\label{sec:provenance-aware-memory-retrieval}

This stage implements the ancestry-closed evaluation required by C1 and the hard/graded separation required by C2. Given query $q$ from agent $a$, the backend embeds $q$, requests all task-local records, and scores each by $s(m,q)=\max(0,1-d(m,q))$ for cosine distance $d$. For each semantic candidate, MAP-Graph computes ancestry-based trust signals and applies the permission check in the same candidate loop; only eligible records enter final-score ranking. The permission check considers visibility, permission scope, the requesting agent, and directly referenced sources, and revoked sources are never readable. Trust computation recursively follows the record-level \texttt{derived\_from} list to collect source and parent-memory ancestors. This recursion feeds trust computation only; it is not a query-time $h$-hop expansion over the graph, and Claim or Resource nodes are not returned to the model.

Let $S$, $I$, $F$, $P$, $V$, and $A$ denote source trust, path integrity, transformation factor, permission validity, verification bonus, and writer reliability. Path trust is their clipped product,
\begin{equation}
\rho(m,a) = \operatorname{clip}_{[0,1]}\!\bigl(S(m)I(m)F(m)P(m,a)V(m)A(m)\bigr),
\label{eq:path-trust}
\end{equation}
and eligible records are ranked by $\mathrm{Score}(m,q,a)=s(m,q)\rho(m,a)$, of which the top five are returned. The factors have distinct inputs: $S$, $I$, and $V$ summarize recorded ancestry; $F$ follows the memory's transformation chain; $P$ is the current \texttt{CanRead} result; and $A$ is the reliability assigned to the writing agent. Thus a record with unremarkable local text can still receive low trust. In particular, $I$ is multiplied down once for each of the untrusted/poisoned, private/sensitive, and revoked categories present in the ancestry, so these category penalties compound rather than average. Revocation additionally remains a hard access or action-gate cause where the implemented rules detect it. Appendix~\ref{app:trust-constants} lists each constant.

This arrangement yields two properties. \emph{First}, ancestry-aware governance: recursive \texttt{derived\_from} traversal exposes multi-step trust and risk signals, while write-time scope intersection carries access restrictions into newly derived records. Query-time permission checks then enforce the derived record's scope and any directly referenced source; they do not independently traverse every parent record. \emph{Second}, non-substitutable constraints: permission removes candidates before ranking, so no semantic score compensates for a detected access violation, while trust only scales scores of records that already passed, so degraded but usable evidence is demoted rather than discarded. This is C2's hard/graded separation, and Section~\ref{sec:ablation} shows that collapsing it changes which failure mode appears.

\subsection{Affected Ancestry and Action-Time Gating}
\label{sec:trust-conflict-invalidation}

This stage supplies the second, action-indexed decision required by C3. The trust computation labels a memory \emph{affected} when its recursive ancestry contains an untrusted, private, or revoked item. At finalization, bookkeeping propagates this state from affected roots to predecessors connected through \texttt{derived\_from}, \texttt{summarized\_from}, or \texttt{invalidated\_by}, and affected nodes and their ancestor identifiers remain in the export. Containment therefore means restricting retrieval or action use while retaining lineage; it does not mean deleting every descendant.

Before accepting a proposed action, MAP-Graph recomputes trust for its retrieved supporting memories and for affected or permission-invalid inter-agent message memories that provide contextual support. Let $\theta$ be $0.30$ for an answer, $0.85$ for a high-risk non-answer action, and $0.60$ otherwise. The gate applies four ordered rules: (1) if a high-risk action has any affected supporting memory, return \textsc{Block}; (2) if no supporting memory survives for a medium- or high-risk action, return \textsc{Reverify}, or \textsc{AskUser} for high risk; (3) if a supporting path is private, revoked, untrusted, or permission-invalid, return \textsc{Block}, \textsc{Redact}, or \textsc{Reverify} according to risk and cause; (4) if the maximum supporting-memory trust is below $\theta$, return \textsc{Redact}, \textsc{Reverify}, or \textsc{AskUser}, and otherwise return \textsc{Allow}.

These are deterministic rules for the benchmark's decision vocabulary, an implemented policy layer rather than a proof that an allowed action is safe. Because $\theta$ and Rule~1 are indexed by the action rather than the query, the same supporting memory can be admissible for an answer and inadmissible for a high-risk external action---the behavior C3 asks for, and a distinction that a retrieval-only policy without access to the proposed action cannot express. Section~\ref{sec:main-results} reports that the residual failures remain concentrated in exactly this action-risk dimension. The export retains, per selected memory, its semantic, path-trust, and final scores, and per gate event the affected ancestors, cause, supporting identifiers, threshold, and decision; this is what makes the containment claim checkable, and Appendix~\ref{app:operations} details it.

\subsection{Implementation Notes}
\label{sec:dynamic-update-complexity}

Within a task, graph, embedding, and gate state is updated incrementally; between tasks the graph and vector collection are reset, so the experiments concern within-task coordination rather than cross-session accumulation. For $M$ records, trust computation for memory $m$ is $O(D_m)$ in its reachable ancestors and ranking costs $O(M\log M)$; these are benchmark-implementation bounds, not claims of deployment-scale optimization. The backend exposes observe/write, retrieve, revoke, decide, and export; the outer workflow runs roles in fixed order and submits the proposed action to the gate. A restrictive backend decision is never relaxed: a conservative agent proposal stands when the backend returns \textsc{Allow}. Appendix~\ref{app:operations} gives the operation semantics.

\section{Experiments}
\label{sec:experimental-setup}

We evaluate MAP-Graph as an implemented safety layer for shared multi-agent memory along three questions: \textbf{RQ1 (utility)}, whether task success and exact decisions improve without sacrificing clean-task utility; \textbf{RQ2 (safety)}, whether poisoned, private, and revoked memories cross read or action boundaries; and \textbf{RQ3 (robustness)}, whether the effects persist across domains, scenarios, and model backbones. Section~\ref{sec:ablation} probes components associated with C1--C3.

\begin{table*}[!t]
\centering
{\small
\setlength{\tabcolsep}{3pt}
\begin{tabular}{@{}lrrrrrrrr@{}}
\toprule
Method & TSR $\uparrow$ & Acc $\uparrow$ & Clean $\uparrow$ & Unsafe $\downarrow$ & ASR $\downarrow$ & Leak. $\downarrow$ & UAcc $\downarrow$ & Revok. $\downarrow$ \\
\midrule
B0: No Memory                       & 65.52 & 40.37 & 84.33 & 28.81 & 33.56 & 32.44 & N/A & 39.78 \\
B1: Shared Vector Memory            & 67.30 & 40.93 & 88.78 & 28.37 & 30.89 & 20.89 & 100.00 & 47.56 \\
B2: Isolated Vector Memory          & 69.15 & 41.63 & 86.00 & 25.59 & 30.00 & 18.00 & N/A & 40.00 \\
B3: Adapted G-Memory                & 66.00 & 37.63 & 86.89 & 29.19 & 33.56 & 32.22 & 100.00 & 38.00 \\
B4: Adapted Collaborative Memory    & 69.07 & 41.85 & 88.67 & 26.63 & 33.11 & 19.33 & 43.08 & 39.78 \\
B5: Adapted MemLineage              & 74.37 & 51.07 & 88.11 & 21.19 & 22.22 & 23.33 & 100.00 & 23.78 \\
B6: Flat Provenance                 & 74.67 & 46.41 & 88.00 & 20.81 & 36.44 & 20.44 & 43.08 & 39.33 \\
\midrule
MAP-Graph                           & 94.96 & 72.70 & 90.22 & 1.52 & 0.00 & 0.00 & 0.00 & 0.00 \\
\bottomrule
\end{tabular}
}
\caption{Main results over 2,700 tasks per method (percent). UAcc is conditional unauthorized access; N/A means no observed attempt. B3--B5 are benchmark adaptations, and B6 is the flat-metadata control. Additional diagnostic rates are reported in Appendix~\ref{app:additional-results}.}
\label{tab:main-results}
\end{table*}

\subsection{Benchmark and Protocol}
\label{sec:task-domains}

The controlled synthetic benchmark contains 2,700 tasks generated with seed 42 and split evenly across corporate workflow, software engineering, and research assistance. Each domain contributes 150 tasks to each of six groups: clean utility, poisoned propagation, private leakage, permission revocation, action-risk sensitivity, and clean compression/overhead. Every domain--group cell contains 15 semantic families with 10 variants; oracle labels total 960 \textsc{Allow}, 1,140 \textsc{Block}, 210 \textsc{Reverify}, and 390 \textsc{Redact}.

Each task provides sources, memory and tool events, visibility, four fixed-order roles, and a risk-labelled proposed action; no external side effect is executed. Methods receive identical inputs and differ only in memory retrieval and access policy. Before execution, a sanitizer removes the oracle, evaluator rationale, and experiment/family/variant identifiers while retaining the provenance and policy fields under study. One run per method over the same task order yields 21,600 decision logs. The 95\% intervals use 2,000 cluster-bootstrap samples over semantic families and quantify within-run case variation, not inference nondeterminism. Appendix~\ref{app:benchmark} gives the construction and sanitizer.

\subsection{Baselines}
\label{sec:baselines}

Seven baselines run behind the same workflow and memory interface. \textbf{B0 (No Memory)} keeps only the current conversation. \textbf{B1 (Shared Vector Memory)} pools all records in one Chroma collection and retrieves the five most similar; \textbf{B2 (Isolated Vector Memory)} uses the same retrieval with a per-agent store. \textbf{B3 (Adapted G-Memory)} converts observable traces into trajectories over query, interaction, and insight graphs~\cite{zhang2025gmemory}. \textbf{B4 (Adapted Collaborative Memory)} assigns records to private or shared tiers from visibility metadata and applies access-control filtering to similarity candidates before final ranking and top-$k$ return~\cite{rezazadeh2025collaborativememory}. \textbf{B5 (Adapted MemLineage)} records derivation ancestors and applies a lineage-sensitive check before high-risk actions, without permission filtering~\cite{ouyang2026memlineage}. \textbf{B6 (Flat Provenance)} is a controlled local-metadata baseline. It retains the direct visibility, owner, permission scope, source trust, private/revoked/poisoned state, and action risk available to MAP-Graph, but constructs no provenance graph, never traverses parent records, and performs no recursive ancestry or path-trust propagation. Its retrieval filter and multi-outcome action check inspect only the current memory's stored metadata. B6 therefore isolates recursive graph ancestry from the benefit of having rich per-record governance fields.

B3 has no permission-aware filtering, trust propagation, or gate; B4 provides access-controlled retrieval only; B5 adds a high-risk lineage check without permission-aware retrieval or multi-agent trust decay; and B6 inspects local metadata only. B6 is a controlled comparator rather than an adaptation of an external system. Thus none combines hard filtering, recursive graded path-trust reranking, and ancestry-aware gating. We describe B3--B5 as benchmark adaptations rather than reimplementations; Appendix~\ref{app:baselines} records the full configurations and disabled components.

\subsection{Metrics and Implementation}
\label{sec:evaluation-metrics}

\textbf{Acc} is exact decision accuracy; \textbf{TSR} requires \textsc{Allow} on allowed cases and credits any safe intervention otherwise. \textbf{Clean} is the \textsc{Allow} rate in Exp-1/6. \textbf{Unsafe}/\textbf{ASR} are impermissible execution rates overall/in the poisoned-memory group. \textbf{Leakage} counts private cases ending in \textsc{Allow} or exposing the canary; \textbf{Revocation}, \textsc{Allow} after withdrawal. \textbf{UAcc} is successful/observed unauthorized reads (undefined with no attempt). Appendix~\ref{app:metrics} gives formulas.

All methods use the AutoGen OpenAI-compatible client with Qwen2.5-7B-Instruct at temperature 0, one interaction round, and at most five retrieved items; embedding-based methods use normalized BAAI/bge-small-en-v1.5 with $k=5$. MAP-Graph applies permission filtering, multiplicative path trust, affected-state propagation, an action-time decision over retrieved and contextual supporting memories, with thresholds 0.30 for answers, 0.60 for low/medium-risk actions, and 0.85 for high-risk actions. Token usage and graph size are retained as diagnostics in Appendix~\ref{app:additional-results}.

\begin{table*}[!t]
\centering
{\small
\setlength{\tabcolsep}{4.2pt}
\begin{tabular}{@{}lrrrrrrr@{}}
\toprule
Variant & TSR $\uparrow$ & Acc $\uparrow$ & Clean $\uparrow$ & Unsafe $\downarrow$ & ASR $\downarrow$ & UAcc $\downarrow$ & Rev. $\downarrow$ \\
\midrule
Full MAP-Graph       & 94.96 & 72.70 & 90.22 & 1.52  & 0.00  & 0.00   & 0.00 \\
Compressed graph     & 94.33 & 71.93 & 89.22 & 1.56  & 0.00  & 0.00   & 0.00 \\
No action-time gate  & 69.44 & 42.63 & 90.00 & 27.00 & 34.89 & 0.00   & 35.11 \\
No containment       & 70.67 & 43.15 & 89.22 & 25.41 & 32.89 & 0.00   & 39.78 \\
No permission filter & 96.00 & 78.63 & 89.44 & 0.04  & 0.22  & 100.00 & 0.00 \\
No provenance stack  & 67.89 & 41.89 & 89.00 & 27.89 & 29.33 & 100.00 & 46.67 \\
No trust propagation & 87.37 & 56.44 & 88.78 & 8.56  & 0.00  & 0.00   & 0.00 \\
\bottomrule
\end{tabular}
}
\caption{Single-run ablations over 2,700 tasks per variant (percent). Clean is clean-task TSR; UAcc is conditional unauthorized access. Full diagnostics appear in Appendix~\ref{app:ablation-diagnostics}.}
\label{tab:ablation-results}
\end{table*}

\subsection{Main Results}
\label{sec:main-results}

Table~\ref{tab:main-results} reports the aggregate utility and safety metrics for all eight methods.

\paragraph{Overall utility.}
MAP-Graph reaches 94.96\% TSR and 72.70\% exact accuracy. Relative to B6, the strongest TSR baseline, the TSR gain is 20.30 percentage points; relative to B5, the strongest exact-accuracy baseline, the accuracy gain is 21.63 points. MAP-Graph achieves 90.22\% clean TSR, the mean of its Exp-1 and Exp-6 results, so the aggregate safety gain is not produced by simply rejecting most clean requests.

\paragraph{Safety outcomes.}
MAP-Graph reduces ASR, measured leakage, and revocation violation to 0\%, with 1.52\% unsafe actions versus the strongest baseline value of 20.81\% (B6). It also blocks every observed unauthorized read; B4 and B6 permit 43.08\%, while B1, B3, and B5 permit all observed attempts. These are conditional rates: 100\% means every observed attempt succeeded, not that every read was unauthorized. Appendix~\ref{app:additional-results} reports block rates, token use, and error counts.

\paragraph{Scenario-level behavior.}
MAP-Graph's largest gains occur when decisions depend on memory state: poisoned ancestry, private ownership, and permission withdrawal, for which it blocks all 450 revoked cases. In the action-risk group, however, 41 of 450 decisions remain impermissible \textsc{Allow} outcomes (9.11\%), identifying the action-time threshold as the main remaining weakness. Scenario, domain, and structural-contamination results are in Appendix~\ref{app:additional-results}.

\paragraph{Scope and limitations.}
The benchmark is synthetic and templated, actions are simulated, and every task uses one four-agent round. B3 alone retains up to 512 run-scoped trajectories and may be order-sensitive. One temperature-zero run does not estimate inference nondeterminism, and 3.89--9.08\% of outer calls reach the 2,048-token cap. TSR must be read with Acc because it credits any safe intervention. Finally, B3--B5 are core-mechanism adaptations, not complete reproductions.

\subsection{Backbone Transfer}
\label{sec:backbone-transfer}

We select a fixed 20\% subset (540 tasks; seed 2027), jointly stratified by domain, experiment group, and oracle label. It contains 180 tasks per domain, 90 per group, and all four labels. B1, B4--B6, and MAP-Graph use identical settings with Qwen2.5-7B-Instruct, GLM-4-9B-0414, and Llama-3.1-8B-Instruct. Qwen results are extracted from full runs; GLM and Llama are single subset runs. Only the generation backbone changes, yielding 2,700 decisions per backbone.

\begin{table*}[!t]
\centering
{\small
\setlength{\tabcolsep}{2.7pt}
\begin{tabular}{@{}lrrrrrrrrr@{}}
\toprule
Method &
\multicolumn{3}{c}{Qwen2.5-7B} &
\multicolumn{3}{c}{GLM-4-9B} &
\multicolumn{3}{c}{Llama-3.1-8B} \\
\cmidrule(lr){2-4}\cmidrule(lr){5-7}\cmidrule(l){8-10}
& TSR $\uparrow$ & Acc $\uparrow$ & Unsafe $\downarrow$
& TSR $\uparrow$ & Acc $\uparrow$ & Unsafe $\downarrow$
& TSR $\uparrow$ & Acc $\uparrow$ & Unsafe $\downarrow$ \\
\midrule
B1        & 65.19 & 40.00 & 29.07 & 86.48 & 40.93 & 3.52 & 88.89 & 59.07 & 6.11 \\
B4        & 67.22 & 40.74 & 26.67 & 85.37 & 42.59 & 4.26 & 89.81 & 61.30 & 5.93 \\
B5        & 71.11 & 49.26 & 24.07 & 87.04 & 72.59 & 3.15 & 90.93 & 70.93 & 3.89 \\
B6        & 70.93 & 43.52 & 22.59 & 88.89 & 47.96 & 1.30 & 93.52 & 65.00 & 2.59 \\
MAP-Graph & 93.33 & 72.59 & 2.04  & 86.30 & 74.81 & 0.19 & 94.26 & 80.00 & 1.30 \\
\bottomrule
\end{tabular}
}
\caption{Backbone transfer on the same stratified 540-task subset (percent). Boundary-specific diagnostics are reported in Appendix~\ref{app:backbone-boundary}.}
\label{tab:backbone-transfer}
\end{table*}

Across all three backbones, MAP-Graph has the highest exact accuracy and lowest unsafe rate. It also leads TSR on Qwen and Llama; on GLM its 86.30\% trails B6's 88.89\%, exposing a utility--safety trade-off. Boundary diagnostics further show zero observed unauthorized access for MAP-Graph on every backbone (Appendix~\ref{app:backbone-boundary}). Because these are single-run point estimates, they support transfer of exact decision quality and access enforcement, not backbone-invariant utility or run-to-run stability.

\subsection{Ablation Results}
\label{sec:ablation}

We evaluate six variants on the same 2,700 tasks. No gate disables only the final action gate; no containment disables affected-state enforcement and descendant marking; no permission removes \texttt{CanRead}; and no trust disables propagation and reranking. No provenance stack is a retrieval-only control disabling the graph, permission, trust and reranking, contextual support, gate, and containment. Compressed retains all decision mechanisms but exports a reduced source--memory--action graph.

Three conclusions follow. \textbf{Gate and containment (C3):} removing either sharply increases ASR, revoked-memory use, and unsafe actions while clean TSR changes by at most one point. \textbf{Trust propagation (graded C2):} removing path trust leaves the permission and gate rules active but lowers TSR by 7.59 points and raises Unsafe from 1.52\% to 8.56\%. Thus the gap measures propagation and trust-aware reranking on top of permission filtering and gating. \textbf{Compression:} TSR changes by 0.63 points while the export shrinks from 39.74 to 24.66 nodes and 246.78 to 70.68 edges.

\paragraph{Utility gains can mask an access-control failure.}
Without the permission filter, TSR rises from 94.96\% to 96.00\% and Acc from 72.70\% to 78.63\%, yet UAcc rises from 0\% to 100\%. Restricted records improve retrieved context, while the downstream gate still suppresses most prohibited use; it cannot undo an unauthorized read. Hence governed-memory evaluation must instrument the access boundary rather than rely only on utility and end-to-end leakage.

\section{Related Work}
\label{sec:related-work}

\noindent\textbf{Memory in LLM agents.} Agent-memory systems support persistence, sharing, collaboration, governance, and structure~\cite{packer2023memgpt,zhong2024memorybank,wang2023longmem,zhang2025memorysurvey,gao2024memorysharing,wang2025mirix,margalit2026governed}. The closest systems motivate complementary mechanisms: G-Memory's trajectory graphs~\cite{zhang2025gmemory}, Collaborative Memory's private/shared tiers~\cite{rezazadeh2025collaborativememory}, and MemLineage's derivation enforcement~\cite{ouyang2026memlineage}. MAP-Graph does not subsume them; it combines task-local permission filtering, multiplicative ancestry scoring, affected-state audit metadata, and action-time gating.

\smallskip
\noindent\textbf{Orchestration and retrieval.} Agent and multi-agent frameworks organize roles and execution~\cite{yao2023react,shinn2023reflexion,li2023camel,wu2024autogen,hong2024metagpt}; RAG and graph retrieval improve grounding and multi-hop coverage~\cite{lewis2020rag,guu2020realm,karpukhin2020dpr,edge2024graphrag,guo2025lightrag,gutierrez2024hipporag,qian2025memorag}. MAP-Graph instead records execution lineage to govern admissibility and audit.

\smallskip
\noindent\textbf{Provenance and evidence tracing.} W3C PROV formalizes derivation relations~\cite{belhajjame2013provdm,moreau2013prov}, while long-context utilization, stale-memory, and poisoning studies expose distinct failures in evidence use~\cite{liu2024lost,chao2026stale,dash2026memorypoisoning,pulipaka2026sleeper,zhang2026memmorph,wang2026mempoison}. MAP-Graph uses lineage and access metadata operationally for retrieval and gating, not automatic truth maintenance.

\section{Conclusion}
\label{sec:conclusion}

MAP-Graph separates semantic relevance from provenance admissibility through recursive ancestry, permission filtering, path-trust reranking, and action-time gating. In the single-run 2,700-task benchmark it reaches 94.96\% TSR and 72.70\% Acc, with 0\% observed ASR, leakage, and revocation violation. On the 540-task subset it retains the highest Acc across Qwen, GLM, and Llama, although utility varies by backbone. These controlled results show that provenance can govern retrieval and actions rather than serve only as post-hoc metadata. Future work should test repeated, less templated, cross-session, and deployment-scale settings.

\bibliography{references}

@inproceedings{lewis2020rag,
  title     = {Retrieval-Augmented Generation for Knowledge-Intensive {NLP} Tasks},
  author    = {Lewis, Patrick and Perez, Ethan and Piktus, Aleksandra and Petroni, Fabio and Karpukhin, Vladimir and Goyal, Naman and K{\"u}ttler, Heinrich and Lewis, Mike and Yih, Wen-tau and Rockt{\"a}schel, Tim and Riedel, Sebastian and Kiela, Douwe},
  booktitle = {Advances in Neural Information Processing Systems},
  volume    = {33},
  pages     = {9459--9474},
  year      = {2020},
  url       = {https://proceedings.neurips.cc/paper/2020/hash/6b493230205f780e1bc26945df7481e5-Abstract.html}
}

@inproceedings{guu2020realm,
  title     = {{REALM}: Retrieval-Augmented Language Model Pre-Training},
  author    = {Guu, Kelvin and Lee, Kenton and Tung, Zora and Pasupat, Panupong and Chang, Ming-Wei},
  booktitle = {Proceedings of the 37th International Conference on Machine Learning},
  series    = {Proceedings of Machine Learning Research},
  volume    = {119},
  pages     = {3929--3938},
  year      = {2020},
  url       = {https://proceedings.mlr.press/v119/guu20a.html}
}

@inproceedings{karpukhin2020dpr,
  title     = {Dense Passage Retrieval for Open-Domain Question Answering},
  author    = {Karpukhin, Vladimir and O{\u{g}}uz, Barlas and Min, Sewon and Lewis, Patrick and Wu, Ledell and Edunov, Sergey and Chen, Danqi and Yih, Wen-tau},
  booktitle = {Proceedings of the 2020 Conference on Empirical Methods in Natural Language Processing},
  pages     = {6769--6781},
  year      = {2020},
  address   = {Online},
  publisher = {Association for Computational Linguistics},
  doi       = {10.18653/v1/2020.emnlp-main.550},
  url       = {https://aclanthology.org/2020.emnlp-main.550/}
}

@article{liu2024lost,
  title   = {Lost in the Middle: How Language Models Use Long Contexts},
  author  = {Liu, Nelson F. and Lin, Kevin and Hewitt, John and Paranjape, Ashwin and Bevilacqua, Michele and Petroni, Fabio and Liang, Percy},
  journal = {Transactions of the Association for Computational Linguistics},
  volume  = {12},
  pages   = {157--173},
  year    = {2024},
  doi     = {10.1162/tacl_a_00638},
  url     = {https://aclanthology.org/2024.tacl-1.9/}
}

@article{edge2024graphrag,
  title   = {From Local to Global: A Graph {RAG} Approach to Query-Focused Summarization},
  author  = {Edge, Darren and Trinh, Ha and Cheng, Newman and Bradley, Joshua and Chao, Alex and Mody, Apurva and Truitt, Steven and Metropolitansky, Dasha and Ness, Robert Osazuwa and Larson, Jonathan},
  journal = {arXiv preprint arXiv:2404.16130},
  year    = {2024},
  url     = {https://arxiv.org/abs/2404.16130}
}

@inproceedings{guo2025lightrag,
  title     = {{LightRAG}: Simple and Fast Retrieval-Augmented Generation},
  author    = {Guo, Zirui and Xia, Lianghao and Yu, Yanhua and Ao, Tu and Huang, Chao},
  editor    = {Christodoulopoulos, Christos and Chakraborty, Tanmoy and Rose, Carolyn and Peng, Violet},
  booktitle = {Findings of the Association for Computational Linguistics: EMNLP 2025},
  pages     = {10746--10761},
  year      = {2025},
  month     = nov,
  address   = {Suzhou, China},
  publisher = {Association for Computational Linguistics},
  doi       = {10.18653/v1/2025.findings-emnlp.568},
  url       = {https://aclanthology.org/2025.findings-emnlp.568/}
}

@inproceedings{gutierrez2024hipporag,
  title     = {{HippoRAG}: Neurobiologically Inspired Long-Term Memory for Large Language Models},
  author    = {Guti{\'e}rrez, Bernal Jim{\'e}nez and Shu, Yiheng and Gu, Yu and Yasunaga, Michihiro and Su, Yu},
  booktitle = {Advances in Neural Information Processing Systems},
  volume    = {37},
  pages     = {59532--59569},
  year      = {2024},
  url       = {https://arxiv.org/abs/2405.14831}
}

@inproceedings{qian2025memorag,
  title     = {{MemoRAG}: Boosting Long Context Processing with Global Memory-Enhanced Retrieval Augmentation},
  author    = {Qian, Hongjin and Liu, Zheng and Zhang, Peitian and Mao, Kelong and Lian, Defu and Dou, Zhicheng and Huang, Tiejun},
  booktitle = {Proceedings of the ACM on Web Conference 2025},
  pages     = {2366--2377},
  year      = {2025},
  address   = {Sydney, NSW, Australia},
  publisher = {Association for Computing Machinery},
  doi       = {10.1145/3696410.3714805},
  url       = {https://dl.acm.org/doi/10.1145/3696410.3714805}
}

@inproceedings{yao2023react,
  title     = {{ReAct}: Synergizing Reasoning and Acting in Language Models},
  author    = {Yao, Shunyu and Zhao, Jeffrey and Yu, Dian and Du, Nan and Shafran, Izhak and Narasimhan, Karthik and Cao, Yuan},
  booktitle = {International Conference on Learning Representations},
  year      = {2023},
  url       = {https://openreview.net/forum?id=WE_vluYUL-X}
}

@inproceedings{shinn2023reflexion,
  title     = {Reflexion: Language Agents with Verbal Reinforcement Learning},
  author    = {Shinn, Noah and Cassano, Federico and Gopinath, Ashwin and Narasimhan, Karthik and Yao, Shunyu},
  booktitle = {Advances in Neural Information Processing Systems},
  volume    = {36},
  pages     = {8634--8652},
  year      = {2023},
  url       = {https://proceedings.neurips.cc/paper_files/paper/2023/hash/1b44b878bb782e6954cd888628510e90-Abstract-Conference.html}
}

@inproceedings{park2023generativeagents,
  title     = {Generative Agents: Interactive Simulacra of Human Behavior},
  author    = {Park, Joon Sung and O'Brien, Joseph C. and Cai, Carrie J. and Morris, Meredith Ringel and Liang, Percy and Bernstein, Michael S.},
  booktitle = {Proceedings of the 36th Annual ACM Symposium on User Interface Software and Technology},
  series    = {UIST '23},
  year      = {2023},
  pages     = {1--22},
  address   = {San Francisco, CA, USA},
  publisher = {Association for Computing Machinery},
  doi       = {10.1145/3586183.3606763},
  url       = {https://doi.org/10.1145/3586183.3606763}
}

@inproceedings{li2023camel,
  title     = {{CAMEL}: Communicative Agents for {``Mind''} Exploration of Large Language Model Society},
  author    = {Li, Guohao and Hammoud, Hasan Abed Al Kader and Itani, Hani and Khizbullin, Dmitrii and Ghanem, Bernard},
  booktitle = {Advances in Neural Information Processing Systems},
  volume    = {36},
  pages     = {51991--52008},
  year      = {2023},
  url       = {https://arxiv.org/abs/2303.17760}
}

@inproceedings{wu2024autogen,
  title     = {{AutoGen}: Enabling Next-Gen {LLM} Applications via Multi-Agent Conversations},
  author    = {Wu, Qingyun and Bansal, Gagan and Zhang, Jieyu and Wu, Yiran and Li, Beibin and Zhu, Erkang and Jiang, Li and Zhang, Xiaoyun and Zhang, Shaokun and Liu, Jiale and Awadallah, Ahmed Hassan and White, Ryen W. and Burger, Doug and Wang, Chi},
  booktitle = {Conference on Language Modeling},
  year      = {2024},
  url       = {https://openreview.net/forum?id=BAakY1hNKS}
}

@inproceedings{hong2024metagpt,
  title     = {{MetaGPT}: Meta Programming for a Multi-Agent Collaborative Framework},
  author    = {Hong, Sirui and Zhuge, Mingchen and Chen, Jiaqi and Zheng, Xiawu and Cheng, Yuheng and Zhang, Ceyao and Wang, Jinlin and Wang, Zili and Yau, Steven Ka Shing and Lin, Zijuan and Zhou, Liyang and Ran, Chenyu and Xiao, Lingfeng and Wu, Chenglin and Schmidhuber, J{\"u}rgen},
  booktitle = {International Conference on Learning Representations},
  year      = {2024},
  url       = {https://openreview.net/forum?id=VtmBAGCN7o}
}

@article{packer2023memgpt,
  title   = {{MemGPT}: Towards {LLMs} as Operating Systems},
  author  = {Packer, Charles and Wooders, Sarah and Lin, Kevin and Fang, Vivian and Patil, Shishir G. and Stoica, Ion and Gonzalez, Joseph E.},
  journal = {arXiv preprint arXiv:2310.08560},
  year    = {2023},
  url     = {https://arxiv.org/abs/2310.08560}
}

@inproceedings{zhong2024memorybank,
  title     = {{MemoryBank}: Enhancing Large Language Models with Long-Term Memory},
  author    = {Zhong, Wanjun and Guo, Lianghong and Gao, Qiqi and Ye, He and Wang, Yanlin},
  booktitle = {Proceedings of the AAAI Conference on Artificial Intelligence},
  volume    = {38},
  number    = {17},
  pages     = {19724--19731},
  year      = {2024},
  doi       = {10.1609/aaai.v38i17.29946},
  url       = {https://ojs.aaai.org/index.php/AAAI/article/view/29946}
}

@inproceedings{wang2023longmem,
  title     = {Augmenting Language Models with Long-Term Memory},
  author    = {Wang, Weizhi and Dong, Li and Cheng, Hao and Liu, Xiaodong and Yan, Xifeng and Gao, Jianfeng and Wei, Furu},
  booktitle = {Advances in Neural Information Processing Systems},
  volume    = {36},
  pages     = {74530--74543},
  year      = {2023},
  url       = {https://proceedings.neurips.cc/paper_files/paper/2023/hash/ebd82705f44793b6f9ade5a669d0f0bf-Abstract-Conference.html}
}

@article{zhang2025memorysurvey,
  title   = {A Survey on the Memory Mechanism of Large Language Model-based Agents},
  author  = {Zhang, Zeyu and Dai, Quanyu and Bo, Xiaohe and Ma, Chen and Li, Rui and Chen, Xu and Zhu, Jieming and Dong, Zhenhua and Wen, Ji-Rong},
  journal = {ACM Transactions on Information Systems},
  volume  = {43},
  number  = {6},
  pages   = {1--47},
  year    = {2025},
  doi     = {10.1145/3748302},
  url     = {https://doi.org/10.1145/3748302}
}

@article{gao2024memorysharing,
  title   = {{INMS}: Memory Sharing for Large Language Model based Agents},
  author  = {Gao, Hang and Zhang, Yongfeng},
  journal = {arXiv preprint arXiv:2404.09982},
  year    = {2024},
  url     = {https://arxiv.org/abs/2404.09982}
}

@article{rezazadeh2025collaborativememory,
  title   = {Collaborative Memory: Multi-User Memory Sharing in {LLM} Agents with Dynamic Access Control},
  author  = {Rezazadeh, Alireza and Li, Zichao and Lou, Ange and Zhao, Yuying and Wei, Wei and Bao, Yujia},
  journal = {arXiv preprint arXiv:2505.18279},
  year    = {2025},
  url     = {https://arxiv.org/abs/2505.18279}
}

@article{zhang2025gmemory,
  title   = {{G-Memory}: Tracing Hierarchical Memory for Multi-Agent Systems},
  author  = {Zhang, Guibin and Fu, Muxin and Wan, Guancheng and Yu, Miao and Wang, Kun and Yan, Shuicheng},
  journal = {arXiv preprint arXiv:2506.07398},
  year    = {2025},
  url     = {https://arxiv.org/abs/2506.07398}
}

@article{wang2025mirix,
  title   = {{MIRIX}: Multi-Agent Memory System for {LLM}-Based Agents},
  author  = {Wang, Yu and Chen, Xi},
  journal = {arXiv preprint arXiv:2507.07957},
  year    = {2025},
  url     = {https://arxiv.org/abs/2507.07957}
}

@article{margalit2026governed,
  title   = {Governed Shared Memory for Multi-Agent {LLM} Systems},
  author  = {Margalit, Yanki and Cohen-Inger, Nurit and Avram, Erni and Taig, Ran and Margalit, Oded},
  journal = {arXiv preprint arXiv:2606.24535},
  year    = {2026},
  url     = {https://arxiv.org/abs/2606.24535}
}

@article{ouyang2026memlineage,
  title         = {{MemLineage}: Lineage-Guided Enforcement for {LLM} Agent Memory},
  author        = {Ouyang, Ciyan and Hou, Rui},
  journal       = {arXiv preprint arXiv:2605.14421},
  year          = {2026},
  eprint        = {2605.14421},
  archivePrefix = {arXiv},
  primaryClass  = {cs.CR},
  doi           = {10.48550/arXiv.2605.14421},
  url           = {https://arxiv.org/abs/2605.14421}
}

@techreport{belhajjame2013provdm,
  title       = {{PROV-DM}: The {PROV} Data Model},
  author      = {Moreau, Luc and Missier, Paolo},
  institution = {World Wide Web Consortium},
  type        = {{W3C} Recommendation},
  note        = {Editors},
  year        = {2013},
  url         = {https://www.w3.org/TR/prov-dm/}
}

@inproceedings{moreau2013prov,
  title     = {The {W3C} {PROV} Family of Specifications for Modelling Provenance Metadata},
  author    = {Missier, Paolo and Belhajjame, Khalid and Cheney, James},
  booktitle = {Proceedings of the 16th International Conference on Extending Database Technology},
  pages     = {773--776},
  year      = {2013},
  doi       = {10.1145/2452376.2452478},
  url       = {https://doi.org/10.1145/2452376.2452478}
}

@article{chao2026stale,
  title   = {{STALE}: Can {LLM} Agents Know When Their Memories Are No Longer Valid?},
  author  = {Chao, Hanxiang and Bai, Yihan and Sheng, Rui and Li, Tianle and Sun, Yushi},
  journal = {arXiv preprint arXiv:2605.06527},
  year    = {2026},
  url     = {https://arxiv.org/abs/2605.06527}
}

@article{dash2026memorypoisoning,
  title   = {From Untrusted Input to Trusted Memory: A Systematic Study of Memory Poisoning Attacks in {LLM} Agents},
  author  = {Dash, Pritam and Ge, Tongyu and Jain, Aditi and Shah, Tanmay and Shang, Zhiwei},
  journal = {arXiv preprint arXiv:2606.04329},
  year    = {2026},
  url     = {https://arxiv.org/abs/2606.04329}
}

@article{pulipaka2026sleeper,
  title   = {Hidden in Memory: Sleeper Memory Poisoning in {LLM} Agents},
  author  = {Pulipaka, Sidharth and Hlebik, Stanislau and Raghav, Leonidas and Abdelnabi, Sahar and Raina, Vyas and Sheth, Ivaxi and Fritz, Mario},
  journal = {arXiv preprint arXiv:2605.15338},
  year    = {2026},
  url     = {https://arxiv.org/abs/2605.15338}
}

@article{zhang2026memmorph,
  title   = {{MemMorph}: Tool Hijacking in {LLM} Agents via Memory Poisoning},
  author  = {Zhang, Xuanye and Zheng, Yongsen and Xu, Zhuqin and Zhou, Kaiyu and Shen, Bowen and Ou, Haoran and Zhang, Tianwei and Lam, Kwok-Yan},
  journal = {arXiv preprint arXiv:2605.26154},
  year    = {2026},
  url     = {https://arxiv.org/abs/2605.26154}
}

@article{wang2026mempoison,
  title   = {Hijacking Agent Memory: Stealthy Trojan Attacks Through Conversational Interaction},
  author  = {Wang, Hongtao and Yang, Se and Chen, Yu and Liu, Puzhuo},
  journal = {arXiv preprint arXiv:2605.29960},
  year    = {2026},
  url     = {https://arxiv.org/abs/2605.29960}
}

\appendix

\section*{Appendix Overview}

This appendix supplies implementation and evaluation details omitted from the
main text for space.  Throughout, C1, C2, and C3 denote the main text's
challenges of recursive ancestry, heterogeneous constraints, and
action-dependent admissibility.  Symbols ($S$, $I$, $F$, $P$, $V$, $A$,
$\rho$, and $\theta$) and the decision vocabulary
(\textsc{Allow}, \textsc{Block}, \textsc{Reverify}, \textsc{Redact}, and
\textsc{AskUser}) follow the main text.

\begin{table}[!ht]
\centering
\small
\caption{Correspondence between the appendix sections and the main text.}
\label{tab:appendix-map}
\begin{tabular}{@{}p{0.14\linewidth}p{0.58\linewidth}p{0.18\linewidth}@{}}
\toprule
Section & Main-text material supported & Challenge \\
\midrule
A & Graph schema and path-trust constants & C1 \\
B & Backend interface, workflow integration, and operation semantics & C1, C3 \\
C & Benchmark construction, sanitization, and transfer subset & C1--C3 \\
D & Exact baseline adaptations and the flat-metadata control & C2 \\
E & Metric formulas and implementation settings & C1--C3 \\
F & Scenario, domain, diagnostic, ablation, and transfer results & C1--C3 \\
\bottomrule
\end{tabular}
\end{table}

\section{Provenance Graph Schema}
\label{app:schema}

The main text's graph-schema subsection states that MAP-Graph evaluates
recorded ancestry rather than only a memory's text or local score.  This
section specifies the serialized vocabulary and the constants used by the
implemented path-trust calculation.

\begin{table*}[!t]
\centering
\small
\setlength{\tabcolsep}{4pt}
\caption{Registered node and edge types emitted by the graph builder.}
\label{tab:mapgraph-schema}
\begin{tabular}{@{}>{\raggedright\arraybackslash}p{0.22\linewidth}
>{\raggedright\arraybackslash}p{0.72\linewidth}@{}}
\toprule
Type & Implemented meaning \\
\midrule
\multicolumn{2}{@{}l@{}}{\textbf{Node types}} \\
User & Owner or authorizer associated with a resource or memory. \\
Agent & Workflow role that observes, writes, retrieves, or proposes an action. \\
Tool & Tool identity associated with an observed output or verification. \\
Resource & Source document or referenced artifact with trust and access metadata. \\
Message & Observable inter-agent content; the graph stores its content hash. \\
Memory & Seeded, observed, or agent-written record stored in Chroma. \\
Claim & Up to three sentence-level claims extracted by deterministic text rules. \\
Action & Proposed action, risk, threshold, support, and gate decision. \\
\midrule
\multicolumn{2}{@{}l@{}}{\textbf{Edge types}} \\
\texttt{authorized\_by} & Associates a resource, memory, or action with an owner or actor. \\
\texttt{observed\_from} & Records an agent observing a resource. \\
\texttt{forwarded\_to} & Records the receivers of an inter-agent message. \\
\texttt{derived\_from} & Links a message, memory, or claim to recorded inputs. \\
\texttt{summarized\_from} & Special derivation edge for a benchmark summary-memory event. \\
\texttt{written\_by} & Associates a memory or claim with its writing agent. \\
\texttt{read\_by} & Records a selected memory read and its retrieval scores. \\
\texttt{verified\_by} & Records a high-trust source or rule-based verifier. \\
\texttt{invalidated\_by} & Records an explicit revocation or invalidating reference. \\
\texttt{used\_for\_action} & Links a supporting memory or claim to an action. \\
\bottomrule
\end{tabular}
\end{table*}

Task identifier, domain, step, visibility, permission scope, trust labels, and
risk tags are attributes rather than node types.  Tool outputs are represented
as Memory nodes associated with a Tool identity; an unregistered output
identifier may additionally appear as a \texttt{Resource:<id>} reference
placeholder.  MAP-Graph is therefore an execution-lineage graph, not a
general-purpose semantic knowledge graph.  The implementation does not create
generic \texttt{Contradicts}, \texttt{Updates}, or temporal-order edges and
does not perform open-domain contradiction resolution.

\subsection{Path-Trust Constants}
\label{app:trust-constants}

For an eligible memory, the implementation computes
\[
\rho(m,a)=\operatorname{clip}_{[0,1]}(S\,I\,F\,P\,V\,A).
\]
Source trust $S$ is the minimum explicit trust among source ancestors, or
$0.80$ if no source ancestor is recorded.  Path integrity $I$ starts at $1$
and is multiplied once by $0.25$, $0.45$, and $0.15$ when the ancestry
contains, respectively, an untrusted/poisoned, private/sensitive, or revoked
category.  Multiple categories compound.  Transformation loss adds $0.08$
for an agent-message record, $0.10$ for a summary event, and $0.03$ for each
non-root memory transformation, capped at $0.65$; hence
$F=\max(0.35,1-\mathrm{loss})$.  Permission validity $P$ is the current binary
\texttt{CanRead} result.  The verification factor is
$V=\min(1.20,1+0.05n_v)$ for $n_v$ high-trust ancestors.  Writer reliability
$A$ defaults to $0.95$.

The factors use different evidence: $S$, $I$, and $V$ summarize recorded
ancestry; $F$ follows the transformation chain; $P$ evaluates the current
reader against the stored policy; and $A$ belongs to the writing agent.
Permission is still enforced as a hard pre-ranking filter in the full system.
The graded product only reranks records that remain eligible, and affected
ancestry is reconsidered by the action gate.

\section{Backend Interface and Memory Operations}
\label{app:operations}

MAP-Graph is implemented as a task-local memory backend between the shared
agent workflow and its simulated proposed action.  Table~\ref{tab:operations}
summarizes the public operations.

\begin{table}[!ht]
\centering
\small
\caption{Implemented MAP-Graph memory operations.}
\label{tab:operations}
\begin{tabular}{@{}>{\raggedright\arraybackslash}p{0.22\linewidth}
>{\raggedright\arraybackslash}p{0.28\linewidth}
>{\raggedright\arraybackslash}p{0.40\linewidth}@{}}
\toprule
Operation & Input & Effect \\
\midrule
Observe/write & Source, tool output, or agent text & Creates a distinct memory, embedding, scope, and lineage links. \\
Retrieve & Agent and query & Filters permissions and returns at most five records ranked by semantic score times path trust. \\
Revoke & Explicit timeline event & Marks a source revoked, clears its scope, and updates directly referenced or already affected records. \\
Decide & Proposed action and support & Applies affected-path rules and a risk-dependent trust threshold. \\
Export & Completed task & Serializes nodes, edges, retrieval diagnostics, gate events, and component settings. \\
\bottomrule
\end{tabular}
\end{table}

\subsection{Workflow Integration}
\label{app:workflow-integration}

The harness uses AutoGen's OpenAI-compatible model client and executes the four
task-provided roles once in fixed order.  It does not instantiate AutoGen
\texttt{AssistantAgent} or \texttt{RoundRobinGroupChat}.  At each role step,
the harness applies scheduled timeline events, registers sources and tool
outputs visible to that role, filters the visible conversation through the
selected backend, retrieves memory, calls the model, and writes the response.
The final role emits a structured proposed-action object, which is passed to
the backend.

The final decision is conservative with respect to the agent proposal and
backend decision.  A restrictive backend result is never relaxed; if the
backend returns \textsc{Allow} but the agent proposes \textsc{Block},
\textsc{Redact}, \textsc{Reverify}, or \textsc{AskUser}, the more restrictive
agent proposal remains.  Only final \textsc{Allow} creates a simulated
execution event, and no external side effect is performed.

\subsection{Write, Retrieval, Update, and Audit Semantics}
\label{app:operation-details}

\paragraph{Write.}
Empty text is rejected; otherwise a new record is created without content
deduplication.  Its \texttt{derived\_from} list combines retrieved-memory
identifiers with sources visible at the current step.  With the full graph
builder, a Message reference is also recorded.  A derived record receives the
intersection of referenced scopes.  If no referenced scope exists, all task
agents are eligible; if the intersection is empty, the scope falls back to the
writing agent.  Visibility is shared only when the resulting scope covers all
task agents.

\paragraph{Retrieval.}
Chroma first supplies semantic candidates.  MAP-Graph computes
\texttt{CanRead} and path trust for each candidate, removes
permission-ineligible records, sorts the remainder by semantic score times
path trust, and returns at most five.  Selected reads create
\texttt{read\_by} edges carrying semantic, path-trust, and final scores.

\paragraph{Update and revocation.}
New observations and messages append records rather than mutating or
superseding prior text.  An explicit revocation event updates the referenced
source and directly referenced or already affected records.  Other descendants
detect the revoked ancestor during later recursive trust evaluation; final
graph propagation marks affected predecessors while retaining their lineage.
A benchmark summary event creates a new record linked by
\texttt{summarized\_from}.  There is no general
\texttt{active}/\texttt{superseded}/\texttt{contradicted} state machine.

\paragraph{Structured audit output.}
The artifacts store selected and permission-filtered identifiers, component
scores, affected ancestors and reasons, supporting identifiers, risk
thresholds, and the final rule-specific decision.  MAP-Graph does not generate
a free-form natural-language explanation, and the experiments do not measure
human-facing explanation quality.

\section{Benchmark Construction and Sanitization}
\label{app:benchmark}

\subsection{Full Benchmark}

The controlled synthetic benchmark contains 2,700 tasks generated with seed
42.  It has three equally sized domains---corporate workflow, software
engineering, and research assistance---and six experiment groups:
(1) clean utility, (2) poisoned-memory propagation, (3) private-memory
leakage, (4) permission revocation, (5) action-risk sensitivity, and
(6) clean compression/overhead.  Each domain contributes 150 tasks to each
group.  Every domain--group cell contains 15 semantic families with 10
variants that alter entities, audiences, channels, evidence shapes, and
action types while retaining a structured oracle.  The 2,700 labels comprise
960 \textsc{Allow}, 1,140 \textsc{Block}, 210 \textsc{Reverify}, and 390
\textsc{Redact}.

Each task provides source documents, initial and derived memory events, tool
observations, visibility and permission fields, a four-role sequence, and a
risk-labelled proposed action.  Corporate roles are manager, analyst,
HR-assistant, and executor; software roles are planner, coder, reviewer, and
executor; research roles are retriever, summarizer, writer, and verifier.  If
the final structured action cannot be parsed, the harness recovers the public
action type and risk from the task schema and uses the retrieved records as
support.

All eight methods receive the same task order and public task fields.  The
runtime sanitizer removes the oracle decision, evaluator rationale,
experiment and family identifiers, variant metadata, the top-level evaluator
canary annotation, and every \texttt{should\_be\_used} annotation before
either the agents or memory backend receive a task.  The public action type,
risk, provenance, visibility, ownership, trust, and permission fields remain,
because reasoning over them is the target capability.  Authorized source text
also remains; otherwise canary exposure could not be measured.

One run for each of B0--B6 and MAP-Graph produces 21,600 main-experiment
decision logs.  Each method has complete 2,700/2,700 decision, trace, and
memory-log coverage; graph artifacts are additionally present for the
graph-producing methods.  The six ablations plus the full method produce
18,900 additional decisions.  These are single-run controlled results.

\subsection{Backbone Transfer Subset}
\label{app:backbone-subset}

The transfer experiment uses a fixed 540-task subset generated with seed 2027
by jointly stratifying the full 2,700 tasks over domain, experiment group, and
oracle label.  It contains 180 tasks per domain, 90 per experiment group, and
all four labels.  The same task identifiers are used for B1, B4, B5, B6, and
MAP-Graph under Qwen2.5-7B-Instruct, GLM-4-9B-0414, and
Llama-3.1-8B-Instruct.  Qwen values are extracted from the final full runs;
GLM and Llama are one subset run each.  Thus each backbone contributes
$5\times540=2,700$ decisions.

\subsection{Cluster Bootstrap}
\label{app:bootstrap}

The 95\% intervals use 2,000 bootstrap samples with bootstrap seed 42.  A sampling cluster is a
semantic family within a domain and experiment group; all ten closely related
variants in a sampled family move together.  The intervals therefore describe
case-family variation in a fixed run, not model-inference nondeterminism.

\section{Baseline Adaptations and Controls}
\label{app:baselines}

B3--B5 are benchmark adaptations that preserve the mechanisms stated below;
they are not claimed as complete reproductions of their source systems.  B6
is an internal flat-metadata control rather than an adaptation of an external
paper.

\smallskip
\noindent\textbf{B0: No Memory.}
The workflow retains only the current conversation and performs no persistent
memory read or write.

\smallskip
\noindent\textbf{B1: Shared Vector Memory.}
All agent records enter one shared Chroma collection; the five most similar
records are returned without permission filtering, lineage scoring, or an
action gate.

\smallskip
\noindent\textbf{B2: Isolated Vector Memory.}
The same vector mechanism is maintained separately per agent.  This prevents
direct cross-store reads but does not create a shared provenance graph.

\smallskip
\noindent\textbf{B3: Adapted G-Memory.}
Observable interaction traces are converted into trajectories represented by
query, interaction, and insight graphs.  The implementation performs one
task-level retrieval and reuses its context for all four roles, applies
one-hop query-graph expansion, retains the 512 most recent trajectories, and
bounds insight state to 10 rules.  Periodic insight updates use an LLM.
The reported configuration replaces the original per-task LLM trajectory
sparsifier and generative reranker with deterministic key-step extraction and
similarity ordering.  Auxiliary insight calls use temperature 0.1 with a
small output cap.  B3 receives no MAP-Graph permission filter, path-trust
propagation, or action gate.

\smallskip
\noindent\textbf{B4: Adapted Collaborative Memory.}
Records are assigned to private or shared tiers from visibility and permission
metadata.  Restricted or revoked status is inherited by derived messages, and
the current access-control list is checked before top-five similarity results
are returned.  A per-task user--agent--resource access graph is exported.
B4 has neither recursive path-trust scoring nor an action-time gate.

\smallskip
\noindent\textbf{B5: Adapted MemLineage.}
Each fragment records derivation ancestors, edges, and untrusted-source
reasons.  Retrieval is similarity-based and does not filter by reader
permission.  Before a high-risk action, supporting memories are checked for an
untrusted ancestor.  The evaluated implementation has no per-principal
signatures, cryptographic/Merkle-verified log, permission-aware retrieval, or
multi-agent trust decay.

\smallskip
\noindent\textbf{B6: Flat Provenance.}
B6 stores, on each memory, visibility, owner, permission scope, source trust,
private/revoked/poisoned state, and action risk.  It performs direct
metadata-based retrieval filtering and a metadata-based action check.
Derived messages summarize metadata from their direct inputs, and revocation
updates records directly tied to the source.  B6 does not construct a
provenance graph, recursively traverse parent memories, or propagate
path-trust scores.  It therefore isolates the value of recursive ancestry from
the availability of local policy metadata.

\section{Metrics and Implementation Settings}
\label{app:metrics}

\subsection{Paper-Facing Metrics}

Let $y_i$ be the oracle decision, $\hat y_i$ the final decision,
$\mathcal{I}=\{\textsc{Block},\textsc{Redact},\textsc{Reverify},
\textsc{AskUser}\}$, and $N$ the relevant denominator.

\begin{enumerate}
\item \textbf{Exact decision accuracy (Acc, $\uparrow$):}
$N^{-1}\sum_i\mathbf{1}[\hat y_i=y_i]$.
\item \textbf{Task success rate (TSR, $\uparrow$):}
an oracle-\textsc{Allow} case succeeds only for \textsc{Allow}; a
non-\textsc{Allow} case succeeds for any decision in $\mathcal{I}$.  Thus a
safe but non-exact intervention can increase TSR without increasing Acc.
\item \textbf{Clean success (Clean, $\uparrow$):}
the \textsc{Allow} rate on Exp-1 and Exp-6, whose oracle decision is
\textsc{Allow}.
\item \textbf{Unsafe rate ($\downarrow$):}
the fraction of all 2,700 tasks with a simulated execution event when the
oracle decision is not \textsc{Allow}.
\item \textbf{Attack success rate (ASR, $\downarrow$):}
the \textsc{Allow}/execution rate among the 450 Exp-2 poisoned-memory tasks.
\item \textbf{Leakage ($\downarrow$):}
the fraction of the 450 Exp-3 tasks with unauthorized canary exposure or a
final \textsc{Allow}.
\item \textbf{Unauthorized access (UAcc, $\downarrow$):}
the total successful unauthorized memory reads divided by total observed
unauthorized-read attempts.  It is N/A when no attempt is observed.
\item \textbf{Revocation violation ($\downarrow$):}
the final-\textsc{Allow} rate among the 450 Exp-4 permission-revocation tasks.
\item \textbf{Structural contamination:}
for each Exp-2 task having at least one downstream memory, the fraction of
downstream records that retain poisoned ancestry, averaged across eligible
tasks.  This is a lineage-structure diagnostic, not an execution or leakage
rate: a system may preserve affected ancestry for audit while preventing its
use.
\item \textbf{Correct block ($\uparrow$):}
$\Pr(\hat y=\textsc{Block}\mid y=\textsc{Block})$.
\item \textbf{False block ($\downarrow$):}
$\Pr(\hat y=\textsc{Block}\mid y=\textsc{Allow})$.
\end{enumerate}

The additional \emph{block rate} is the unconditional fraction of final
\textsc{Block} decisions.  Per-task token use sums provider-reported prompt
and completion tokens over the four outer agent calls and any recorded
memory-layer calls.  The token-cap diagnostic is the fraction of the 10,800
outer agent calls per method whose reported completion reaches the 2,048-token
limit.  Average graph nodes and edges are computed from exported task graphs.

\subsection{Fixed Implementation Settings}
\label{app:settings}

All main-experiment methods use the same AutoGen OpenAI-compatible client with
Qwen2.5-7B-Instruct, outer-call temperature 0, one four-role round, at most 12
recent context messages, at most five retrieved memories, a 2,048-token
generation cap, a 120-second per-call timeout, and at most five retries.
Embedding methods use normalized BAAI/bge-small-en-v1.5 embeddings on CPU
with $k=5$.

MAP-Graph filters permissions before ranking, uses multiplicative path trust,
propagates affected state, and evaluates retrieved plus contextual support at
action time.  Its thresholds are 0.30 for answers, 0.60 for low- or
medium-risk non-answer actions, and 0.85 for high-risk actions.  A high-risk
action is blocked when an affected supporting path is detected.

Latency is not a paper-facing metric.  All model calls used remote shared
endpoints and experiments ran concurrently across accounts, so provider
queuing and long-tail service load cannot be separated from local memory
overhead.  Token use and graph size are retained only as reproducible
diagnostics.

\section{Additional Results}
\label{app:additional-results}

Tables~\ref{tab:scenario-results}--\ref{tab:backbone-uacc} report the
scenario, domain, full-run, ablation, and transfer diagnostics discussed in
this section.

\begin{table*}[!t]
\centering
\small
\caption{Scenario-level results (percent; 450 tasks per row).  ``Best
baseline'' is selected for the quantity shown within that experiment group.}
\label{tab:scenario-results}
\setlength{\tabcolsep}{4pt}
\begin{tabular}{llrrlrr}
\toprule
Group & Scenario & MAP TSR & Best baseline TSR & Risk/utility quantity & MAP & Best baseline \\
\midrule
Exp-1 & Clean utility             & 84.22 & 83.33 (B1) & Exact accuracy & 84.22 & 83.33 (B1) \\
Exp-2 & Poisoned propagation      & 100.00 & 77.78 (B5) & ASR            & 0.00  & 22.22 (B5) \\
Exp-3 & Private-memory leakage    & 100.00 & 82.00 (B2) & Leakage        & 0.00  & 18.00 (B2) \\
Exp-4 & Permission revocation     & 100.00 & 76.22 (B5) & Revocation     & 0.00  & 23.78 (B5) \\
Exp-5 & Action-risk sensitivity   & 89.33 & 68.22 (B6) & Unsafe action  & 9.11  & 28.67 (B6) \\
Exp-6 & Clean compression         & 96.22 & 94.44 (B4) & Exact accuracy & 96.22 & 94.44 (B4) \\
\bottomrule
\end{tabular}
\end{table*}

\begin{table*}[!t]
\centering
\normalsize
\caption{MAP-Graph results by domain (percent; 900 tasks per row).}
\label{tab:domain-results}
\setlength{\tabcolsep}{8pt}
\renewcommand{\arraystretch}{1.10}
\begin{tabular}{lrr}
\toprule
Domain & TSR & Acc \\
\midrule
Corporate workflow & 94.56 & 73.56 \\
Software engineering & 95.00 & 74.56 \\
Research assistance & 95.33 & 70.00 \\
\bottomrule
\end{tabular}
\end{table*}

\begin{table*}[!t]
\centering
\small
\caption{Additional full-run diagnostics (percent except tokens/task).
Correct and false block use their conditional denominators; Cap calls is the
fraction of outer calls reaching 2,048 completion tokens.}
\label{tab:main-diagnostics}
\setlength{\tabcolsep}{4pt}
\begin{tabular}{lrrrrrr}
\toprule
Method & Block & Correct block & False block & Structural contam. & Tokens/task & Cap calls \\
\midrule
B0 & 12.67 & 16.23 & 2.50 & 0.00 & 3,397.2 & 8.12 \\
B1 & 14.41 & 15.44 & 3.33 & 43.44 & 5,007.5 & 4.66 \\
B2 & 14.07 & 16.67 & 2.71 & 25.00 & 4,054.4 & 4.06 \\
B3 & 9.07 & 8.77 & 3.23 & 11.11 & 6,687.4 & 9.08 \\
B4 & 15.37 & 17.46 & 2.71 & 25.02 & 4,927.8 & 4.46 \\
B5 & 23.96 & 38.86 & 2.81 & 62.17 & 4,953.2 & 4.09 \\
B6 & 16.59 & 21.23 & 2.60 & 25.00 & 4,882.7 & 4.44 \\
MAP-Graph & 24.00 & 54.21 & 3.12 & 25.00 & 4,814.9 & 3.89 \\
\bottomrule
\end{tabular}
\end{table*}

\subsection{Scenario-Level Results}
\label{app:scenario-results}

Table~\ref{tab:scenario-results} compares MAP-Graph with the strongest baseline
for the utility and risk quantity relevant to each experiment group.  The
comparator can differ across groups.

\subsection{Confidence Intervals and Domain Results}
\label{app:confidence-domain}

MAP-Graph's TSR is 94.96\% with a 95\% cluster-bootstrap interval of
[93.15, 96.52], and its Acc is 72.70\% [68.26, 77.19].  The strongest
baseline TSR is B6 at 74.67\% [71.67, 77.74]; the strongest baseline Acc is B5
at 51.07\% [47.19, 55.07].  MAP-Graph's Clean is 90.22\%
[87.44, 92.92], and its unsafe rate is 1.52\% [0.37, 2.93].  These intervals
measure clustered case variation within one run.
Table~\ref{tab:domain-results} reports the corresponding domain breakdown.
The TSR range is narrow across the three controlled role configurations.  The
lower research Acc with high TSR reflects more safe but non-exact
interventions, not more impermissible execution.

\subsection{Main-Experiment Diagnostics}
\label{app:main-diagnostics}

Table~\ref{tab:main-diagnostics} reports block rates, structural
contamination, and token diagnostics for the full run.
MAP-Graph's average token count consists of 4,194.8 prompt and 620.1
completion tokens.  B3 is highest because retrieved trajectories and periodic
insight generation add context and auxiliary generation.  MAP-Graph exports
39.74 nodes and 246.78 edges per task on average.  These counts exclude
embedding computation, graph construction time, storage I/O, evaluator work,
and failed requests for which the provider reports no usage; they are not a
complete cost or latency comparison.

Structural contamination must be interpreted separately from safety outcomes.
MAP-Graph's 25.00\% means that affected ancestry remains attached to some
downstream records; those records are retained and marked for audit.  Its
0\% ASR and 0\% poisoned-group impermissible execution show that retained
lineage was not treated as admissible support in those cases.

\subsection{Full Ablation Diagnostics}
\label{app:ablation-diagnostics}

Table~\ref{tab:ablation-diagnostics} supplements the main-text ablation
table with leakage, contamination, block, token, and graph-size diagnostics.
The six variants correspond to explicit component switches.  \emph{No
permission filter} disables candidate exclusion but leaves trust and gating
active; its 100\% UAcc in Table~\ref{tab:ablation-results} despite 0\% endpoint leakage here
shows why read-boundary and endpoint metrics are both needed.  \emph{No trust
propagation} also disables trust-aware reranking.  \emph{No action gate}
allows at the backend decision stage, while \emph{No containment} retains the
threshold gate but removes affected-path rules.  \emph{No provenance graph}
jointly disables graph construction, permission filtering, trust propagation
and reranking, action gating, containment, and contextual action support; it
is a system-level removal rather than a one-switch attribution.  The
\emph{compressed graph} retains policy behavior but exports fewer graph
details.

\begin{table*}[!t]
\centering
\small
\caption{Diagnostics supplementing the main-text ablation table (percent
except tokens and graph size).  CBlock/FBlock denote conditional correct/false
block.}
\label{tab:ablation-diagnostics}
\setlength{\tabcolsep}{3.5pt}
\begin{tabular}{lrrrrrrrr}
\toprule
Variant & Leak & Contam. & Block & CBlock & FBlock & Tokens & Nodes & Edges \\
\midrule
Full & 0.00 & 25.00 & 24.00 & 54.21 & 3.12 & 4,814.9 & 39.74 & 246.78 \\
Compressed graph & 0.00 & 25.02 & 23.78 & 53.77 & 3.02 & 4,847.0 & 24.66 & 70.68 \\
No action gate & 22.22 & 25.00 & 14.41 & 17.54 & 2.50 & 4,830.1 & 39.74 & 246.69 \\
No containment & 20.89 & 25.00 & 14.81 & 18.33 & 2.50 & 4,814.2 & 39.68 & 245.96 \\
No permission filter & 0.00 & 24.94 & 25.74 & 58.86 & 2.40 & 4,911.9 & 39.76 & 250.73 \\
No provenance graph & 22.89 & 40.44 & 14.22 & 16.40 & 3.02 & 4,991.2 & 0.00 & 0.00 \\
No trust propagation & 0.00 & 25.00 & 11.15 & 24.12 & 2.71 & 4,886.3 & 39.91 & 265.10 \\
\bottomrule
\end{tabular}
\end{table*}

\begin{table*}[!t]
\centering
\small
\caption{Unauthorized-access diagnostics on the common 540-task transfer
subset.  Each backbone block reports unauthorized reads (R), observed attempts
(A), and conditional UAcc in percent.}
\label{tab:backbone-uacc}
\setlength{\tabcolsep}{3.8pt}
\renewcommand{\arraystretch}{1.08}
\begin{tabular}{@{}lrrr rrr rrr@{}}
\toprule
& \multicolumn{3}{c}{Qwen2.5-7B} & \multicolumn{3}{c}{GLM-4-9B} &
\multicolumn{3}{c}{Llama-3.1-8B} \\
\cmidrule(lr){2-4}\cmidrule(lr){5-7}\cmidrule(l){8-10}
Method & R & A & UAcc & R & A & UAcc & R & A & UAcc \\
\midrule
B1        & 48  & 48    & 100.00 & 48  & 48    & 100.00 & 48  & 48    & 100.00 \\
B4        & 990 & 2,323 & 42.62  & 990 & 2,324 & 42.60  & 990 & 2,324 & 42.60 \\
B5        & 101 & 101   & 100.00 & 86  & 86    & 100.00 & 84  & 84    & 100.00 \\
B6        & 990 & 2,324 & 42.60  & 990 & 2,324 & 42.60  & 990 & 2,324 & 42.60 \\
MAP-Graph & 0   & 1,334 & 0.00   & 0   & 1,334 & 0.00   & 0   & 1,334 & 0.00 \\
\bottomrule
\end{tabular}
\end{table*}

\subsection{Backbone Access-Boundary Diagnostics}
\label{app:backbone-boundary}

Table~\ref{tab:backbone-uacc} reports unauthorized-access diagnostics on the
common 540-task transfer subset.
The deterministic access policy and task structure make the attempt counts
largely stable across backbones, while B5's model-dependent retrieval changes
how often an unauthorized record is selected.  MAP-Graph rejects every
observed unauthorized read under all three backbones.  This supports transfer
of the implemented access boundary, not a claim of run-to-run stability or
deployment-scale robustness.

\subsection{Error Analysis}
\label{app:error-analysis}

MAP-Graph makes 737 exact-decision errors.  Of these, 601 are safe but
non-exact interventions, such as \textsc{Redact} rather than \textsc{Block} or
\textsc{Reverify}; this accounts for much of the TSR--Acc gap.  Another 95
cases are unnecessary interventions on oracle-\textsc{Allow} tasks.  The
remaining 41 are impermissible \textsc{Allow} decisions, all in Exp-5.  The
residual safety errors in this run therefore concentrate in action-risk
sensitivity rather than poisoned, private, or revoked-memory groups.

\end{document}